\documentclass{article}
\usepackage{times}
\usepackage{helvet}
\usepackage{courier}
\usepackage[english]{babel}
\usepackage[latin1]{inputenc}
\usepackage{amsmath}
\usepackage{color}
\usepackage{amssymb}
\usepackage{amsfonts}
\usepackage{amsthm}
\usepackage{mathrsfs}
\usepackage[round]{natbib}
\usepackage{hyperref}
\title{Agent-Mediated Social Choice}
\author{Umberto Grandi\\ 
Institut de Recherche en Informatique de Toulouse (IRIT) \\University of Toulouse\\
\url{umberto.grandi@ut-capitole.fr}}
\date{}

\begin{document}

\maketitle

\begin{abstract}
\noindent
Direct democracy is often proposed as a possible solution to the 21st-century problems of democracy.
However, this suggestion clashes with the size and complexity of 21st-century societies, entailing an excessive cognitive burden on voters, who would have to submit informed opinions on an excessive number of issues.
In this paper I argue for the development of ``voting avatars'', autonomous agents debating and voting on behalf of each citizen.
Theoretical research from artificial intelligence, and in particular multiagent systems and computational social choice, proposes 21st-century techniques for this purpose, from the compact representation of a voter's preferences and values, to the development of voting procedures for autonomous agents use only.
\end{abstract}

\section{Introduction}\label{sec:intro}

Computational studies of voting are mostly motivated by two intended applications:
the coordination of societies of artificial agents, and the study of human collective decisions whose complexity requires the use of computational techniques.
Both research directions are too often confined to theoretical studies, with unrealistic assumptions constraining their significance for real-world situations.
Most practical applications of these results are therefore confined to low-stakes decisions, which are of great importance in expanding the use of algorithms in society, but are far from high-stakes choices such as political elections, referenda, or parliamentary decisions, which societies still make using old-fashioned technologies like paper ballots.

In this paper I argue in favour of conceiving ``voting avatars'', artificial agents that are able to act as proxies for voters in collective decisions at any level of society. 
Besides being an ideal test-bed for a large number of techniques developed in the field of multiagent systems and artificial intelligence in general, agent-mediated social choice may also suggest innovative solutions to low voter participation, a problem that is endemic in most practical implementations of electronic decision processes.

\paragraph{From low-stakes/high-frequency to high-stakes/high-frequency social choice.}
In their quest for practical applications, researchers in (computational) social choice have argued in favour of moving away from political elections, and high-stakes/low-frequency collective decisions in general, where computational techniques and studies are less relevant.
Low-stakes decisions such as answering to a personalised search query, or designing group recommender systems for retailers, were instead identified as more suitable applications for their research \citep{BoutilierLu2011}. 
A similar trend can be observed in existing platforms for electronic democracy such as
LiquidFeedback\footnote{\url{http://liquidfeedback.org/}}, which were initially designed to be used for policy design by political parties, and evolved into decision systems for smaller committees and low-stakes decisions \citep{BehrensEtAl2014}.
Many researchers, however, (including the author of this paper) became interested in social choice with the dream of having an impact on how large societies take high-stakes decisions.
Technological advancements now make it possible to carry out polls and surveys among citizens in almost real-time.
\emph{Has the time come for high-stakes/high-frequency collective decision-making?}

 \paragraph{Electronic, direct, participative, and interactive democracy.}
Electronic democracy (or e-democracy) is an umbrella term that groups several pieces of software and policies that aim at increasing citizens' participation in collective decisions by means of digital technologies.  
Among those, the most relevant applications for social choice theorists revolve around the development of interactive or direct forms of democracy through the design of electronic platforms.
While on the one hand existing platforms tend to be rather simple, or, as mentioned in the previous paragraphs, they are restricted to low-stakes decisions, on the other hand more encompassing visions of direct democracy often seem excessively utopian in imagining an active participation of the entire electorate to collective debates and votes \citep{Green-Armytage2015}.
Real-world experiments seem to suggest the opposite: when the citizens of Madrid were asked to vote directly (and electronically) on the renovation projects of one of the most important squares in the city, only 8\% of eligible voters actually took part in the voting process.\footnote{\url{https://elpais.com/elpais/2017/02/28/inenglish/1488280371_057827.html}}
The use of direct (or interactive) democracy platforms for high-stakes decisions seems to produce a very low citizen participation, a phenomenon that might be considered as an instance of the well-known paradox of voting \citep{Downs1957}, i.e., the cost of casting a single vote exceeding the expected benefit of affecting the result of the election.
\emph{Can we imagine a technological advancement that will solve the problem of low participation in direct democratic decisions?}

\paragraph{Artificial intelligence and democracy.} Techniques from artificial intelligence (AI), such as the use of machine learning for user profiling and micro-targeting, have been widely used in recent political campaigns, and their effects have been widely debated by the press. 
However, most articles limit the future use of AI in elections to the design of centralised algorithms that take collective decisions from the collection of citizens' preferences and behaviours.\footnote{A vision already proposed by \citet{Asimov} in the short story ``Franchise''.} 
While discussing the potential misuse of data analysis techniques by central governments, a recent article by \citet{ScientificAmerican} claims that ``If data filters and recommendation and search algorithms would be selectable and configurable by the user, we could look at problems from multiple perspectives, and we would be less prone to manipulation by distorted information''.
\emph{Can we develop personalised AI techniques that help people construct and motivate their views and voting behaviour in collective decisions?}

\paragraph{Agent-mediated e-commerce and computational mechanism design.} 
Negotiation technology, trust-building, and a vast number of other techniques developed in multi-agent systems found application in e-commerce, where human and artificial agents participate with various roles in suitably designed markets. 
A research field that started more than fifteen years ago \citep{SierraJAAMAS2004,feigenbaum09,DashEtAlIEEE2003} helped creating a new reality: ``The anticipated agent-mediated economy is almost upon us" \citep{ParkesAAMAS2017}.
\emph{Once people get used to delegate their consumer power to artificial agents, will they be ready to delegate their citizen power as well?}

\paragraph{Computational social choice and its applications.} 
The field of computational social choice (COMSOC) started around ten years ago when a group of researchers with common interests in computer science, economics, and political science regrouped in an
international workshop,\footnote{\url{www.illc.uva.nl/COMSOC/workshops.html}} and flourished up to the recent publication of a handbook \citep{HBCOMSOC2016}. 
Research in this field is rather theoretical, and its motivation often ambiguous between developing technologies for the coordination of artificial agents and the study of theoretical properties of voting mechanisms to be later used by humans. 
Unrealistic assumptions of complete knowledge or full rationality of individual agents limit the applicability of many of its results, a problem that would be less relevant in agent-mediated institutions.
If the field is to succeed and prosper, it needs applications to feed its research agenda, experimenting with either real-world situations \citep[as argued by][]{BouveretTrends} or agent-based technologies. 
A recent positive example is the creation of the Spliddit webpage \citep{ShahAAMAS2017}, which created a fruitful loop between practical applications and theoretical research.
\emph{Can we conceive of technologies enabling agent-mediated social choices, where theorems and algorithms from COMSOC research could be applied without changing their currently unrealistic assumptions?}

\section{Voting Avatars - A Short Story}

Sylvia (a human being) lives in a world where each citizen with the right to vote is paired with a \emph{voting avatar}, an autonomous agent that is able to communicate with her and who is authenticated by a central voting authority to vote on her behalf.\footnote{This section may read like a science-fiction novel, but so do many academic papers advocating for direct democracy.}
Sylvia is the only authorised person to communicate with her voting avatar (e.g., speech recognition, fingerprint authentication, ...). 
The central voting authority is a democratically elected government with executive power, supported by an elected bureaucracy who sets the agenda of debates, polls, and votes to be conducted among the entire electorate.
Each morning Sylvia receives the daily political agenda, with issues classified by themes and by interest: local, regional, national, and global (the world in which Sylvia lives is likely to be a world federation). 
Sylvia can simply ignore the message and have a good cup of coffee, as she does on most days: Her voting avatar has already been searching the internet for opinions, consulted influential avatars, and built a preliminary voting behaviour for her daily agenda.
During the day, the voting avatar will follow all discussions and correlated votes, and update Sylvia's voting behaviour based on this information and on the level of strategic behaviour she set (currently she left the ``strategic voting" button unchecked, like most of her friends claim to do).
Her voting avatar has been training for years on a number of votes Sylvia takes directly every month, as well as by observing her conversations on social media, by reading her emails, and by having direct conversations with her every time that the avatar made a wrong or debatable decision on her behalf.
Last night, for instance, Sylvia realised that the avatar suggested voting against issuing extra visas to refugees from the Mars colonies, based on a number of dubious sources that she had consulted a couple of days earlier out of curiosity. 
They discussed the issue for a good 5 minutes, clarifying her position on immigration, the job market, and charity (she actually found the discussion very helpful in constructing a solid view on these issues).
Today Sylvia is quite interested in the debate on global freezing, and her voting avatar is proposing a vote in support of the current bill (decisions with long-term consequences involve a long series of iterated votes on improving proposals, in order to maximise consensus). Sylvia has access to a short summary of the reasons supporting the avatar's suggestions, with links to a number of articles by authors she finds reliable, extracts from email discussions she had with a friend on this topic, as well as a list of her past decisions on related issues. 
She notices that the coherence warning is yellow, suggesting that her vote clashes with some of the positions she defended on the energy market a couple of months ago, but not being a public figure she chooses to ignore the warning...


\section{Multiagent Systems and Artificial Intelligence}\label{sec:techniques}

In the past 20 years researchers in the field of multiagent systems, a research community in artificial intelligence, have been importing and adapting models from theoretical economics and political science to conceive and program societies of artificial agents \citep[see, e.g., ][]{ShohamLeytonBrown2009}.
Social choice mechanisms have been proposed for collective decision-making in multiagent systems, stimulating research, e.g., on the computational properties of voting rules, the development of tractable procedures for strategic voting, or the development of approximation algorithms for computing the result of particularly hard rules.

Together with algorithmic game theory, computational social choice is currently one of the most well-represented research areas in conferences on multiagent systems.\footnote{See, e.g., the proceedings of the \emph{International Conference on Autonomous Agents and Multiagent Systems, AAMAS} (\url{www.ifaamas.org}).} 
Many of the techniques introduced by COMSOC researchers would find a prime application in the development of voting avatars for agent-mediated collective decisions, such as those described in the short story above. 

\paragraph{Strategic voting.} 
Modelling human voters as perfectly rational agents is a useful simplification for obtaining intuitive theorems, but these assumptions limit significantly the applicability of these results (take the Gibbard-Sattertwaithe Theorem as a classical example). 
Modelling bounded rationality is a big challenge in both AI and Economics, a problem that is absent, or less severe, in societies of artificial agents.
Moreover, many of the techniques developed by COMSOC researchers for the analysis of strategic voting \citep[see, e.g., ][]{ConitzerWalshHandbook2016,FaliszewskiRotheHandbook2016} may find application test-beds in agent-mediated collective choices. 

\paragraph{Machine learning.} 
A voting avatar needs to be able to learn a voter's views and preferences from a set of voters' choices, that can moreover be evolving over time, and be more or less consistent with a set of existing views already present in society.
This certainly is a challenging problem to frame, and one for which large personalised datasets are hard to construct. 
The use of machine learning techniques is still widely unexplored in social choice, and existing papers are mostly focused on creating novel voting methods \citep[see, e.g., ][]{XiaAAMAS2013}. A fruitful starting point may be the conception of decision-support systems based on past voting behaviour.

\paragraph{Iterative voting.} 
Voting methods that have been discarded as too complex or too unintuitive for human voters could be used successfully by artificial agents that act as proxies for voters.
One example is iterative voting, in which repeated elections are staged in search for a more consensual voting outcome \citep[see, e.g.,][for a survey]{MeirTrends}.
The use of voting avatars takes the burden of intensive communication away from the voter, and reinforcement learning techniques could be tested in this setting to obtain novel voting strategies and rules \citep{AiriauEtAlADT2017}.

\paragraph{Combinatorial voting and judgment aggregation.}
By making high-frequency collective decisions possible, the development of voting avatars will be confronted with a large and interconnected space of alternative choices.
This would require compact representations for voters' views and preferences, such as those developed in the area of combinatorial voting \citep{LangXiaHandbook2016}, as well as a detailed understanding of the 
counterintuitive results and paradoxical situations that arise when aggregating them, such as those settings analysed by the theory of judgment aggregation \citep{EndrissHandbook2016}.

\paragraph{Argumentation.} 
Political views and voters' behaviour are the result of discussions and debates, both in the social neighbourhood of a voter and in society in general. The field of (computational) argumentation is a vast enterprise ranging from the elicitation of arguments from natural language, to the identification of winning arguments in a debate between multiple agents. Theoretical research in this field \citep[see, e.g., ][]{RahwanSimari2009} is recently being complemented by technologies that can form the basis of personal assistants for the construction of political views.

\paragraph{Social influence.}
A voting avatar will be scouting for opinions presented in influential newspapers or expressed by a designated set of influential voting avatars. Computational-friendly models of influence need to be developed, perhaps based on the extensive research on trust and reputation in multiagent systems \citep{SabaterSierra2005}. The relation between social choices and social networks have recently been investigated by a number of papers \citep[see, e.g., ][for a survey]{GrandiTrends}


\section{Conclusions and Challenges}\label{sec:conclusions}

We may be quite far from a society in which collective choices of all sorts are taken every day in large numbers by means of artificial agents that vote on our behalf. 
However, in this paper I argue that the implementation of similar ``voting avatars'' is a viable solution to the lack of voter participation which affects e-democracy applications, and would moreover constitute a prime application for many of the techniques developed in computational social choice and multiagent systems in general.
A first step in this direction could be the development of decision-aid systems, which help voters construct their political views and opinions on current issues, building on existing (simple) technologies such as Vote Match\footnote{\url{www.votematch.org.uk}} in UK, or Stemwijzer\footnote{\url{www.stemwijzer.nl}} in the Netherlands.

There are, however, a number of scientific challenges that need to be tackled. 
First, representing a voter's view may require combining techniques from computational knowledge representation with complex models of voting behaviour coming from political science and sociology.
Second, data protection and security will be a key aspect of developing artificial agents acting on behalf of humans. However, this problem is shared by virtually all electronic applications of voting. 
A related problem is vote buying and vote influence in general, given the availability of data about voters' views and votes that might be generated by moving to electronic platforms for collective decision-making. Solutions may be sought in cryptography algorithms, which may be easier to apply to societies of artificial agents rather than human ones (see, e.g., the recent work of \citet{ParkesEtAlAAMAS2017}).
Third, a functional voting avatar needs to be able to explain its behaviour to the voter in an understandable and convincing way. Explanation and interpretation are among the most important challenges for the deployment of artificial intelligence techniques in society, and a recent survey argues in favour of importing sociological theories of explanation to create transparent and trustworthy algorithms \citep{Miller2017}. 
Investigating how an artificial agent can explain simple voting behaviours in terms of the input preferences received may constitute a first step in this direction.
Last, the design of voting avatars that can act as proxies for voters raises all sorts of ethical questions -- from the level of autonomy of such an artificial proxy to the legal status of agent-mediated collective decisions -- some of which are currently debated as general ethical issues related to the employment of artificial intelligence technologies in our societies. 
It turns out that voting itself might be a possible mean of developing collective views on similar ethical problems \citep{NoothigattuEtAlAAI2018}.

\section*{Acknolwedgements}
I am grateful to the participants and the speakers of the reading group on e-democracy at the University of Toulouse.\footnote{\url{www.irit.fr/~Umberto.Grandi/teaching/directdem}} Many thanks to Sylvie Doutre, Piotr Faliszewski, Davide Grossi, Arianna Novaro, Ashley Piggins, Ariel Procaccia, and Marija Slavkovik for their useful comments on previous versions of this paper. 

\bibliographystyle{abbrvnat}
\bibliography{eco-future}

\begin{thebibliography}{28}
\providecommand{\natexlab}[1]{#1}
\providecommand{\url}[1]{\texttt{#1}}
\expandafter\ifx\csname urlstyle\endcsname\relax
  \providecommand{\doi}[1]{doi: #1}\else
  \providecommand{\doi}{doi: \begingroup \urlstyle{rm}\Url}\fi

\bibitem[Airiau et~al.(2017)Airiau, Grandi, and Perotto]{AiriauEtAlADT2017}
S.~Airiau, U.~Grandi, and F.~S. Perotto.
\newblock Learning agents for iterative voting.
\newblock In \emph{Proceedings of the 5th International Conference on
  Algorithmic Decision Theory (ADT)}, 2017.

\bibitem[Asimov(1955)]{Asimov}
I.~Asimov.
\newblock Franchise.
\newblock \emph{If}, 1955.

\bibitem[Behrens et~al.(2014)Behrens, Kistner, Nitsche, and
  Swierczek]{BehrensEtAl2014}
J.~Behrens, A.~Kistner, A.~Nitsche, and B.~Swierczek.
\newblock \emph{Principles of Liquid Feedback}.
\newblock 2014.

\bibitem[Boutilier and Lu(2011)]{BoutilierLu2011}
C.~Boutilier and T.~Lu.
\newblock Probabilistic and utility-theoretic models in social choice:
  Challenges for learning, elicitation, and manipulation.
\newblock In \emph{Proceedings of the IJCAI-2011 Workshop on Social Choice and
  Artificial Intelligence}, 2011.

\bibitem[Bouveret(2017)]{BouveretTrends}
S.~Bouveret.
\newblock Social choice and the web.
\newblock In U.~Endriss, editor, \emph{Trends in Computational Social Choice}.
  AI Access, 2017.

\bibitem[Brandt et~al.(2016)Brandt, Conitzer, Endriss, Lang, and
  Procaccia]{HBCOMSOC2016}
F.~Brandt, V.~Conitzer, U.~Endriss, J.~Lang, and A.~D. Procaccia, editors.
\newblock \emph{Handbook of Computational Social Choice}.
\newblock Cambridge University Press, 2016.

\bibitem[Conitzer and Walsh(2016)]{ConitzerWalshHandbook2016}
V.~Conitzer and T.~Walsh.
\newblock Barriers to manipulation in voting.
\newblock In F.~Brandt, V.~Conitzer, U.~Endriss, J.~Lang, and A.~D. Procaccia,
  editors, \emph{Handbook of Computational Social Choice}. Cambridge University
  Press, 2016.

\bibitem[Dash et~al.(2003)Dash, Jennings, and Parkes]{DashEtAlIEEE2003}
R.~K. Dash, N.~R. Jennings, and D.~C. Parkes.
\newblock Computational-mechanism design: {A} call to arms.
\newblock \emph{{IEEE} Intelligent Systems}, 18\penalty0 (6):\penalty0 40--47,
  2003.

\bibitem[Downs(1957)]{Downs1957}
A.~Downs.
\newblock \emph{An Economic Theory of Democracy}.
\newblock Harper, 1957.

\bibitem[Endriss(2016)]{EndrissHandbook2016}
U.~Endriss.
\newblock Judgment aggregation.
\newblock In F.~Brandt, V.~Conitzer, U.~Endriss, J.~Lang, and A.~D. Procaccia,
  editors, \emph{Handbook of Computational Social Choice}. Cambridge University
  Press, 2016.

\bibitem[Faliszewski and Rothe(2016)]{FaliszewskiRotheHandbook2016}
P.~Faliszewski and J.~Rothe.
\newblock Control and bribery in voting.
\newblock In F.~Brandt, V.~Conitzer, U.~Endriss, J.~Lang, and A.~D. Procaccia,
  editors, \emph{Handbook of Computational Social Choice}. Cambridge University
  Press, 2016.

\bibitem[Feigenbaum et~al.(2009)Feigenbaum, Parkes, and Pennock]{feigenbaum09}
J.~Feigenbaum, D.~C. Parkes, and D.~M. Pennock.
\newblock Computational challenges in e-commerce.
\newblock \emph{Communications of the ACM}, 52:\penalty0 70{\textendash}74,
  2009.

\bibitem[Freedman et~al.(2018)Freedman, Borg, Sinnott-Armstrong, Dickerson, and
  Conitzer]{FreedmanEtAlAAI2018}
R.~Freedman, J.~S. Borg, W.~Sinnott-Armstrong, J.~Dickerson, and V.~Conitzer.
\newblock Adapting a kidney exchange algorithm to align with human values.
\newblock In \emph{Proceedings of the 32nd AAAI Conference on Artificial
  Intelligence (AAAI)}, 2018.

\bibitem[Grandi(2017)]{GrandiTrends}
U.~Grandi.
\newblock Social choice on social networks.
\newblock In U.~Endriss, editor, \emph{Trends in Computational Social Choice}.
  AI Access, 2017.

\bibitem[Green-Armytage(2015)]{Green-Armytage2015}
J.~Green-Armytage.
\newblock Direct voting and proxy voting.
\newblock \emph{Constitutional Political Economy}, 26\penalty0 (2):\penalty0
  190--220, 2015.

\bibitem[Helbing et~al.(February 2017)Helbing, Frey, Gigerenzer, Hafen, Hagner,
  Hofstetter, van~den Hoven, Zicari, and Zwitter]{ScientificAmerican}
D.~Helbing, B.~S. Frey, G.~Gigerenzer, E.~Hafen, M.~Hagner, Y.~Hofstetter,
  J.~van~den Hoven, R.~V. Zicari, and A.~Zwitter.
\newblock Will democracy survive big data and artificial intelligence.
\newblock \emph{Scientific American}, February 2017.

\bibitem[Lang and Xia(2016)]{LangXiaHandbook2016}
J.~Lang and L.~Xia.
\newblock Voting in combinatorial domains.
\newblock In F.~Brandt, V.~Conitzer, U.~Endriss, J.~Lang, and A.~D. Procaccia,
  editors, \emph{Handbook of Computational Social Choice}. Cambridge University
  Press, 2016.

\bibitem[Meir(2017)]{MeirTrends}
R.~Meir.
\newblock Iterative voting.
\newblock In U.~Endriss, editor, \emph{Trends in Computational Social Choice}.
  AI Access, 2017.

\bibitem[Miller(2017)]{Miller2017}
T.~Miller.
\newblock Explanation in artificial intelligence: Insights from the social
  sciences.
\newblock \emph{CoRR}, abs/1706.07269, 2017.
\newblock URL \url{http://arxiv.org/abs/1706.07269}.

\bibitem[Noothigattu et~al.(2018)Noothigattu, Gaikwad, Awad, Dsouza, Rahwan,
  Ravikumar, and Procaccia.]{NoothigattuEtAlAAI2018}
R.~Noothigattu, S.~N.~S. Gaikwad, E.~Awad, S.~Dsouza, I.~Rahwan, P.~Ravikumar,
  and A.~D. Procaccia.
\newblock A voting-based system for ethical decision making.
\newblock In \emph{Proceedings of the 32nd AAAI Conference on Artificial
  Intelligence (AAAI)}, 2018.

\bibitem[Parkes(2017)]{ParkesAAMAS2017}
D.~C. Parkes.
\newblock On {AI}, markets and machine learning.
\newblock In \emph{Proceedings of the 16th Conference on Autonomous Agents and
  MultiAgent Systems (AAMAS)}, 2017.

\bibitem[Parkes et~al.(2017)Parkes, Tylkin, and Xia]{ParkesEtAlAAMAS2017}
D.~C. Parkes, P.~Tylkin, and L.~Xia.
\newblock Thwarting vote buying through decoy ballots.
\newblock In \emph{Proceedings of the 16th Conference on Autonomous Agents and
  MultiAgent Systems, (AAMAS)}, 2017.

\bibitem[Rahwan and Simari(2009)]{RahwanSimari2009}
I.~Rahwan and G.~R. Simari.
\newblock \emph{Argumentation in Artificial Intelligence}.
\newblock Springer, 2009.

\bibitem[Sabater and Sierra(2005)]{SabaterSierra2005}
J.~Sabater and C.~Sierra.
\newblock Review on computational trust and reputation models.
\newblock \emph{Artificial Intelligence Review}, 24\penalty0 (1):\penalty0
  33--60, Sep 2005.

\bibitem[Shah(2017)]{ShahAAMAS2017}
N.~Shah.
\newblock Optimal social decision making.
\newblock In \emph{Proceedings of the 16th International Conference on
  Autonomous Agents and MultiAgent Systems (AAMAS)}, 2017.

\bibitem[Shoham and Leyton-Brown(2009)]{ShohamLeytonBrown2009}
Y.~Shoham and K.~Leyton-Brown.
\newblock \emph{Multiagent Systems: {A}lgorithmic, Game-Theoretic, and Logical
  Foundations}.
\newblock Cambridge University Press, 2009.

\bibitem[Sierra(2004)]{SierraJAAMAS2004}
C.~Sierra.
\newblock Agent-mediated electronic commerce.
\newblock \emph{Autonomous Agents and Multi-Agent Systems}, 9\penalty0
  (3):\penalty0 285--301, 2004.

\bibitem[Xia(2013)]{XiaAAMAS2013}
L.~Xia.
\newblock Designing social choice mechanisms using machine learning.
\newblock In \emph{Proceedings of the 12th International Conference on
  Autonomous Agents and Multi-Agent Systems (AAMAS)}, 2013.

\end{thebibliography}

\end{document}